\def\year{2020}\relax
\documentclass[letterpaper]{article} 
\usepackage{aaai20}  
\usepackage{times}  
\usepackage{helvet} 
\usepackage{courier}  
\usepackage[hyphens]{url}  
\usepackage{graphicx} 
\usepackage{graphicx}  
\usepackage{booktabs}
\usepackage{multirow}
\usepackage{amsmath}
\usepackage{subfigure}
\usepackage{amsfonts,amssymb}
\usepackage{color}
\usepackage{bbding}
\title{Pre-Trained and Attention-Based Neural Networks for Building Noetic Task-Oriented Dialogue Systems}

\author{Jia-Chen Gu$^1$\thanks{Equal contribution.}, Tianda Li$^2$\footnotemark[1], Quan Liu$^1$, Xiaodan Zhu$^2$, Zhen-Hua Ling$^1$, Yu-Ping Ruan$^1$ \\
  $^1$National Engineering Laboratory for Speech and Language Information Processing, \\
      University of Science and Technology of China, Hefei, China \\
  $^2$ECE \& Ingenuity Labs, Queen's University, Kingston, Canada \\
{\tt \{gujc,ypruan\}@mail.ustc.edu.cn}, {\tt \{tianda.li,xiaodan.zhu\}@queensu.ca}, \\ {\tt \{quanliu,zhling\}@ustc.edu.cn}
}

\begin{document}

\maketitle

\begin{abstract}
  The NOESIS II challenge, as the Track 2 of the 8th Dialogue System Technology Challenges (DSTC 8), is the extension of DSTC 7.
  This track incorporates new elements that are vital for the creation of a deployed task-oriented dialogue system.
  This paper describes our systems that are evaluated on all subtasks under this challenge.
  We study the problem of employing pre-trained attention-based network for multi-turn dialogue systems.
  Meanwhile, several adaptation methods are proposed to adapt the pre-trained language models for multi-turn dialogue systems, in order to keep the intrinsic property of dialogue systems.
  In the released evaluation results of Track 2 of DSTC 8, our proposed models ranked fourth in subtask 1, third in subtask 2, and first in subtask 3 and subtask 4 respectively.
\end{abstract}

\section{Introduction}

  Building on the success of Track 1 of the 7th Dialogue System Technology Challenges (DSTC 7) (NOESIS: Noetic End-to-End Response Selection Challenge) \cite{gunasekara2019dstc7}, an extension as Track 2 of DSTC 8 (NOESIS II: Predicting Responses, Identifying Success, and Managing Complexity in Task-Oriented Dialogue) is proposed \cite{DSTC8}, which incorporates new elements that are vital for the creation of a deployed task-oriented dialogue system.
  Specifically, three new dimensions are added to the challenge:
  (1) conversations with more than 2 participants,
  (2) predicting whether a dialogue has solved the problem yet,
  and (3) handling multiple simultaneous conversations.
  Each of these adds an exciting new dimension and brings the task closer to the creation of systems able to handle the complexity of real-world conversation.

  Enabling dialogue systems to converse naturally with humans is a challenging yet intriguing problem of artificial intelligence.
  Recently, human-computer conversation has attracted increasing attention due to its promising potentials and alluring commercial values.
  Dialogue systems aim to engage users in human-computer conversations in the open domain.
  The existing approaches to dialogue systems includes generation-based methods \cite{DBLP:conf/acl/ShangLL15,DBLP:conf/aaai/SerbanSBCP16} and retrieval-based methods \cite{DBLP:conf/sigdial/LowePSP15,DBLP:journals/dad/LowePSCLP17,DBLP:conf/acl/WuWXZL17,DBLP:conf/coling/ZhangLZZL18}. Generation-based models maximize the probability of generating a response given the previous dialogue.
  This approach enables the incorporation of rich context when mapping between consecutive dialogue turns.
  Retrieval-based methods select a proper response for the current conversation from a repository with response selection algorithms, and have the advantage of producing informative and fluent responses.

  Pre-trained language models are also a very popular topic in the domain of natural language processing \cite{DBLP:conf/naacl/PetersNIGCLZ18,DBLP:conf/naacl/DevlinCLT19,DBLP:journals/corr/abs-1906-08237}.
  It captures rich language information from texts and shows great performance on many downstream tasks, such as natural language inference (NLI) \cite{DBLP:conf/emnlp/BowmanAPM15} and question answering (QA) \cite{DBLP:conf/emnlp/RajpurkarZLL16}.
  Typically, the premise in NLI or question in QA is considered as the sentence A, and the hypothesis in NLI or answer in QA is considered as the sentence B.
  A long sequence is formed by concatenating the two sentences with a segmentation token, and then is sent into the model for classification.

  In this paper, we describe our systems that are evaluated on all subtasks of Track 2 of DSTC 8.
  Pre-trained language models are employed to establish pre-trained attention-based models for multi-turn dialogue systems.
  Bidirectional Encoder Representations from Transformers (BERT) \cite{DBLP:conf/naacl/DevlinCLT19}, as one of the most recently developed models, is adopted as the basic of our work.
  Furthermore, to make pre-trained language models more suitable for dialogue systems, several adaptation methods are proposed to keep its intrinsic property.

  As shown in the released evaluation results, our proposed models ranked fourth in subtask 1, third in subtask 2, and first in subtask 3 and subtask 4 respectively.
  In the following sections, we first introduce the related work on dialogues, task descriptions of Track 2 in DSTC 8, and present the details of our proposed model.
  Then the experimental settings and evaluation results are shown.
  Furthermore, experimental results are analyzed by ablation tests.
  Finally we draw conclusions and give an overview of our future work.

  \begin{table*}[t]
    \small
    \centering
    \setlength{\tabcolsep}{0.4mm}{
    \begin{tabular}{clcc}
    \toprule
    Subtask  &  Description   & Ubuntu & Advising \\
    \midrule
       1     &  A conversational context and 100 utterances that could be the next message (either 99 or 100 will be incorrect). & \CheckmarkBold & \CheckmarkBold \\
       2     &  A conversational context contains multiple entangled conversations (either 99 or 100 will be incorrect).         & \CheckmarkBold & \XSolidBrush   \\
       3     &  Participants predict where in a dialogue the problem is solved (if at all). & \XSolidBrush   & \CheckmarkBold \\
       4     &  Given a section of the chat logs, one needs to identify a set of conversations contained within that coherent section. & \CheckmarkBold & \XSolidBrush   \\
    \bottomrule
    \end{tabular}}
    \caption{Task description of each subtask in Track 2.}
    \label{tab1}
  \end{table*}

\section{Related Work}

  The existing methods used to build an open domain dialogue system can be generally categorized into generation-based methods and retrieval-based methods.
  The generation-based models synthesize a response with a natural language generation model by maximizing its generation probability given the previous conversation context.
  This approach enables the incorporation of rich context when mapping between consecutive dialogue turns \cite{DBLP:conf/acl/ShangLL15,DBLP:conf/aaai/SerbanSBCP16}.

  The first two subtasks of Track 2 belong to the retrieval-based task, which learn a matching model for a pair of a conversational context and a response candidate.
  This approach has the advantage of providing informative and fluent responses because they select a proper response for the current conversation from a repository by means of response selection algorithms \cite{DBLP:conf/sigdial/LowePSP15,DBLP:journals/dad/LowePSCLP17,DBLP:conf/acl/WuWXZL17,DBLP:conf/coling/ZhangLZZL18,gu2019utterance}.
  Previous work on retrieval-based chatbots focused on single-turn response selection \cite{DBLP:conf/emnlp/WangLLC13,DBLP:journals/corr/JiLL14}.
  Recently, researchers have extended the focus to the multi-turn conversation, which is more practical for real applications.
  Some earlier work on multi-turn response selection matched a response with concatenating the context utterances literally into a single long sequence, and calculating its matching score with a response candidate \cite{DBLP:conf/sigdial/LowePSP15,DBLP:journals/dad/LowePSCLP17}.
  Recent work has kept utterances separate and performed matching within a representation-interaction-aggregation framework, which improved the performance on this task.
  For example, 
  \cite{DBLP:conf/acl/WuWXZL17} proposed the sequential matching network (SMN) which first matched the response with each utterance and then accumulated the matching information by recurrent neural network (RNN).
  \cite{DBLP:conf/coling/ZhangLZZL18} proposed the deep utterance aggregation network (DUA) which refined utterances and employed self-matching attention to route the vital information in each utterance.
  \cite{DBLP:conf/acl/WuLCZDYZL18} proposed the deep attention matching network (DAM) which constructed representations at different granularities with stacked self-attention and cross-attention.
  \cite{DBLP:conf/wsdm/TaoWXHZY19} proposed the multi-representation fusion network (MRFN) with multiple types of representations.
  \cite{Gu:2019:IMN:3357384.3358140} proposed the interactive matching network (IMN) which performed the global and bidirectional interactions between the context and response.
  \cite{DBLP:conf/acl/TaoWXHZY19} proposed the interaction over interaction (IOI) model which performed matching by stacking multiple interaction blocks.
  \cite{gu-etal-2019-dually} proposed the dually interactive matching network (DIM), which adopted a dual matching architecture by performing the interactive matching between responses and contexts and between responses and personas respectively for ranking response candidates.


  The fourth subtasks of Track 2 can be categorized as disentangle problem.
  Simultaneous conversation occurs not only in informal social interactions but also in multi-party involved chat in our daily life.
  Aiming to separate intermingled messages to detached conversations, disentanglement is of vital importance for understanding conversation.
  The research for conversation disentanglement could date back to \cite{DBLP:journals/corr/abs-cs-0608083} which conducted a study of voice conversations among 8-10 people with an average of 1.76 conversations active at a time.
  Then followed by more research not only propose datasets \cite{mehri2017chat,riou:hal-01698147,kummerfeld2019large} but also models \cite{mehri2017chat,jiang2018learning}.

  However, these models mentioned above were all RNN-based, CNN-based or Transformer-based models without employing any pre-trained language models.
  This paper makes the attempt to employ the pre-trained language model for multi-turn dialogues, and propose new approach for it to keep the intrinsic property of multi-turn dialogue systems.

\section{Task Description}

  The Track 2 of DSTC 8 focuses on task-oriented multi-turn dialogues.
  It is divided into four different subtasks and explores three dialogue challenges: next utterance selection, task success, and conversation disentanglement.
  Two datasets are provided, i.e., Ubuntu and Advising, which will be introduced in detail in the experiment section.
  The series of subtasks has similar structures, but varies in the output space and available context.
  Detailed descriptions of each subtask are shown in Table~\ref{tab1}.
  \CheckmarkBold indicates that the task is evaluated on the marked dataset, and \XSolidBrush indicates not applicable.

\section{Methodology}

  We present here our proposed methods and the detailed implementation.
  Due to limit space, we omit an exhaustive background description of the model architecture of BERT and its basic block Transformer.
  Readers can refer to \cite{DBLP:conf/naacl/DevlinCLT19} and \cite{DBLP:conf/nips/VaswaniSPUJGKP17} for details.

  \subsection{Subtask 1}

  \subsubsection{Input Representation}

    To represent a pair of sentence A and sentence B, the original BERT concatenates this pair of sentence with a \texttt{[SEP]} token.
    For a given token, its input representation of the original BERT is constructed by summing the corresponding token, segment and position embeddings.

    When constructing the sentence A for multi-turn response selection, in order to distinguish the utterances in a context and to model the speaker exchange in turn as the conversation progresses, we use two methods as follows.

\begin{figure}
      \centering
      \includegraphics[width=8cm]{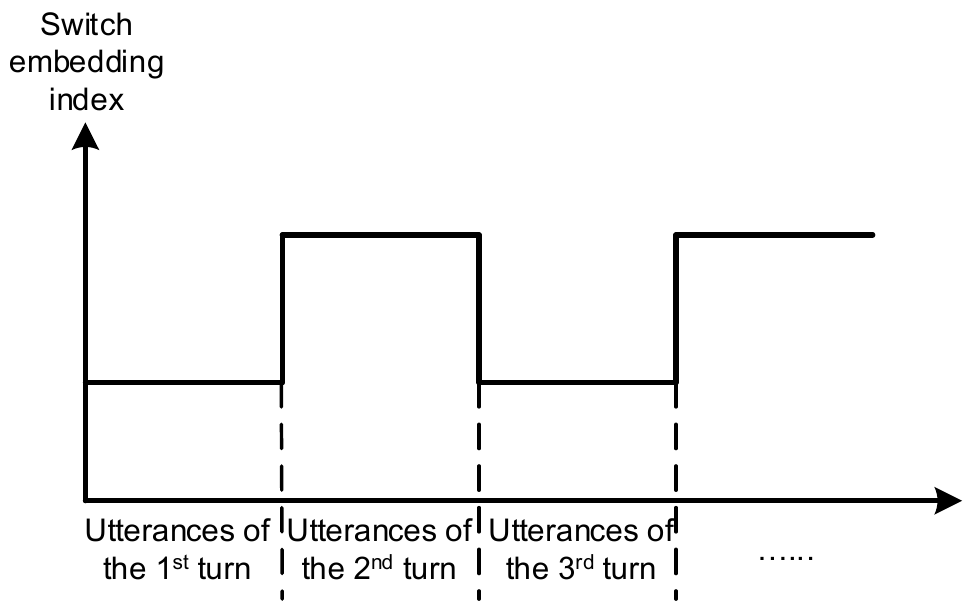}
      \caption{The switch embeddings function as a switch to change the speaker in turn as the conversation progresses.}
      \label{fig2}
    \end{figure}

    \begin{itemize}
      \item \textbf{Segmentation tokens}.
            Empirical results in \cite{DBLP:journals/corr/abs-1802-02614} show that segmentation tokens play an important role for multi-turn response selection.
            Motivated by it, a \texttt{[EOU]} token is added at the end of an utterance and a \texttt{[EOT]} token is added at the end of a turn.
            These tokens can help to model the interactions between utterances in the context implicitly, without using extra complicated networks.
      \item \textbf{Switch embeddings}.
            In order to model the speaker exchange during the conversation directly, we add additional \emph{switch embeddings} to the corresponding token.
            These embeddings function as a switch to change the speaker in turn as the conversation progresses.
            Furthermore, we propose an assumption that conversations are conducted between two speakers, i.e., only two switch embedding vectors are required to be estimated during the training process.
            In this assumption, the first vector is added to the utterances of the first conversation turn.
            When the speaker changes, the second vector is added to the utterances of the second conversation turn.
            Then, the first one is employed again when it comes to the third conversation turn, and so on.
            An illustration of how switch embeddings work is shown in Figure~\ref{fig2}.
            The switch embeddings are expected to model the speaker exchange during the conversation directly to keep the intrinsic property of dialogue systems.
    \end{itemize}

\begin{figure*}
      \centering
      \includegraphics[width=17cm]{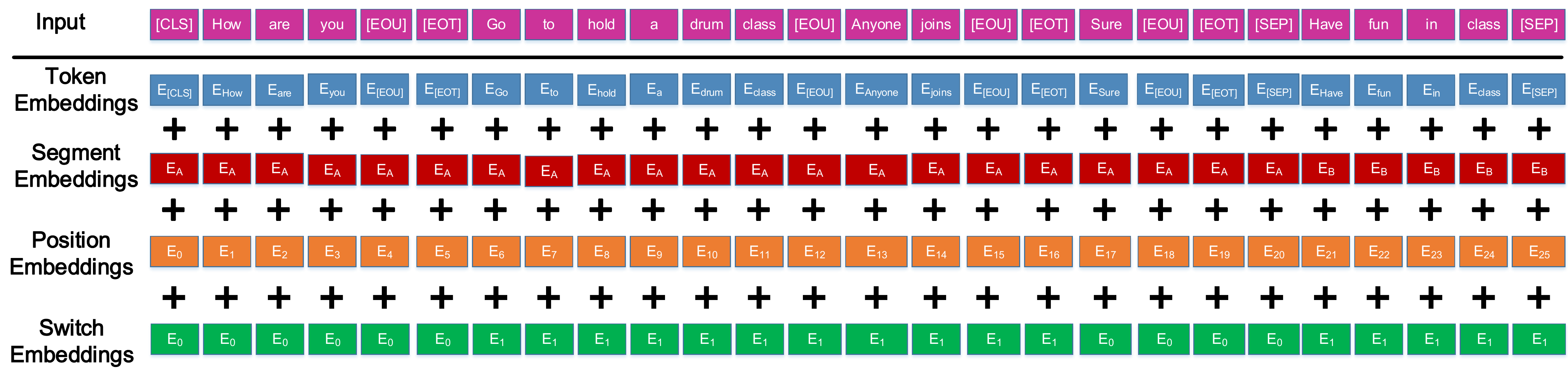}
      \caption{Input representation. The input embeddings is the sum of the token embeddings, the segmentation embeddings, the position embeddings and the switch embeddings.}
      \label{fig1}
    \end{figure*}

    Finally, a visual architecture of our input representation is illustrated in Figure~\ref{fig1}.

  \subsubsection{Output Representation}
    Similar to the original BERT, the first token of each sequence which is the concatenation of a context-response pair is the \texttt{[CLS]} token, whose embedding is used as the aggregated representation for classifying a context-response pair.
    This embedding captures the matching information between a context-response pair, denoted as $\textbf{c} \in \mathbb{R}^H$.
    Then, this embedding is sent into a classifier with a sigmoid output layer as follows:
    \begin{equation}
      s = \textbf{sigmoid}(\textbf{c} \cdot \textbf{W}^\top + \textbf{b}),
    \end{equation}
    where $\textbf{W} \in \mathbb{R}^{1\times H}$ and $\textbf{b} \in \mathbb{R}$ need to be estimated during the fine-tuning process.

    Finally, the classifier returns a score $s$ to denote the matching degree of a context-response pair.

  \subsubsection{Dynamic Negative Sampling}
    When constructing the training set, the positive and negative responses are sampled in a ratio of 1:1.
    For those examples without positive response, the positive one is neglected.
    When sampling the negative responses, we select different negative samples at different epochs.
    Thus, given a context, we fix the positive response and select different negative responses at different epochs so that the model could have a strong ability to distinguish the positive from the negative.

  \subsubsection{Pre-training Tasks}
    The original BERT is trained on a large text corpus to learn general language representations.
    To incorporate some specific in-domain knowledge into language representation models, some pre-training tasks are designed.
    Here, the masked language model (MLM) and the next sentence prediction (NSP) \cite{DBLP:conf/naacl/DevlinCLT19} are employed.
    In addition to the provided dataset for the specific subtask, DSTC 8 provides external files, which contain the source data of both Ubuntu and Advising domains.
    We use the provided external data to pre-train our BERT model, in order to further improve the performance of our model.

    \begin{itemize}
      \item \textbf{MLM}.
            We follow the experimental settings in the original BERT by masking some percentage of the input tokens at random and then predicting only those masked tokens to train a deep bidirectional representation.
            In more detail, we replace the word with the \texttt{[MASK]} token at 80\% of the time, with a random word at 10\% of the time, and with the original word at 10\% of the time.
      \item \textbf{NSP}.
            If there is no pre-training process, the switch embeddings have to be initialized at random at the beginning of the fine-tuning process.
            To achieve a better performance, the switch embeddings can be pre-trained with the help of NSP.
            Here, the sentence A and sentence B are constructed with the same method as we mentioned above.
            The positive responses are true responses that follow the context, and the negative responses are randomly sampled.
            For the Ubuntu dataset, we used title and question as sentence A, and answer as sentence B.
            For the Advising dataset, we use the name of courses as sentence A, and its description as sentence B.
            For every sentence A, we randomly pick another answer as negative sample.
            The embedding of the \texttt{[CLS]} token is used as the aggregated representation for classification.
    \end{itemize}

  \subsection{Subtask 2}
    Subtask 2 is similar to subtask 1 but need additional disentangle strategy.
    When a group of people communicate in a common channel there are often multiple conversation topics occurring concurrently.
    In terms of a specific conversation topic, utterances relevant to it are useful and other utterances could be considered as noise for them.
    Meanwhile, BERT is not good at dealing with sequences which are composed of thousands of tokens as the maximum length of position embeddings is set to 512.
    In order to select a small number of most important utterances, a disentanglement strategy is necessary.

    In this paper, we propose a heuristic speaker-aware strategy to select utterances according to the utterance speakers as follows:
    \begin{itemize}
      \item First, we define the speaker who is uttering an utterance as the \emph{spoken-from} speaker, and define the speaker who is receiving an utterance as the \emph{spoken-to} speaker.
          Each utterance usually has both the \emph{spoken-from} and \emph{spoken-to} speakers.
          But some utterances may have only the \emph{spoken-from} speaker and the \emph{spoken-to} speaker is unknown and considered as \emph{None}.
      \item Second, given the speaker of the response, we select the utterances which have the same \emph{spoken-from} or \emph{spoken-to} speaker as that of the response.
      \item Third, these selected utterances are then organized in their original chronological order and used to form the context.
      \item Finally, utterances with the \emph{spoken-from} or \emph{spoken-to} speaker correspond to the two types of switch embedding respectively.
    \end{itemize}

  \subsection{Subtask 3}
    Subtask 3 is different from the first 2 subtasks, which aims to predict whether and where a dialogue has solved the problem.
    In more detail, the dialogue is conducted between a advisor and a student, in order to help the student select appropriate courses.
    One of \emph{Accept}, \emph{Reject} or \emph{No Decision Yet}, should be made for each utterance.

    We formalize this problem as a combination of sequence labeling and natural language inference.
    Each utterance is considered as a unit when performing sequence labeling.
    Furthermore, a three-class classification of \emph{Accept} (Entailment), \emph{Reject} (Contradiction) and \emph{No Decision Yet} (Neutral) is performed for each unit.
    A hierarchical RNN-based model is adopted rather than the pre-trained language model because the former shows a better performance on this task.

    First, each utterance is encoded by a BiLSTM \cite{DBLP:journals/neco/HochreiterS97} separately at the utterance-level.
    A pooling operation with a combination of max pooling and last-hidden-state pooling is performed to obtain a sequence of utterance embeddings.
    Then, to incorporate the context information, another context-level BiLSTM is employed by considering each utterance as a unit and organizing them in their original chronological order.
    Finally, the outputs of the context-level BiLSTM at each time step are used as the inputs of a multi-layer perceptron (MLP) classifier for classification.

    There are several challenges to this task.
    One is the sparsity of labels because most of utterances belong to the class of \emph{No Decision Yet}.
    To address this problem, we design a weighted loss function that pay larger weights to the \emph{Accept} or \emph{Reject}, which enforces the model to pay more attention to utterances with \emph{Accept} or \emph{Reject} labels.
    Another challenge is the lack of training examples because the training set is small which is composed of only 500 examples.
    We augment the training set by generating some paraphrase examples with the set provided by the track organizer, so that the model could see more different contexts.

  \subsection{Subtask 4}

    Subtask 4 is another disentanglement task that we need to identify several sets of conversations occurring in the same section of IRC channel.
    Specifically, every section contains more than 1000 messages including directed messages posted by users and information messages.
    It contains more than 500 links, and each link indicates that the two linked messages are in the same conversation in time order.

\begin{figure}
      \includegraphics[width=8.5cm]{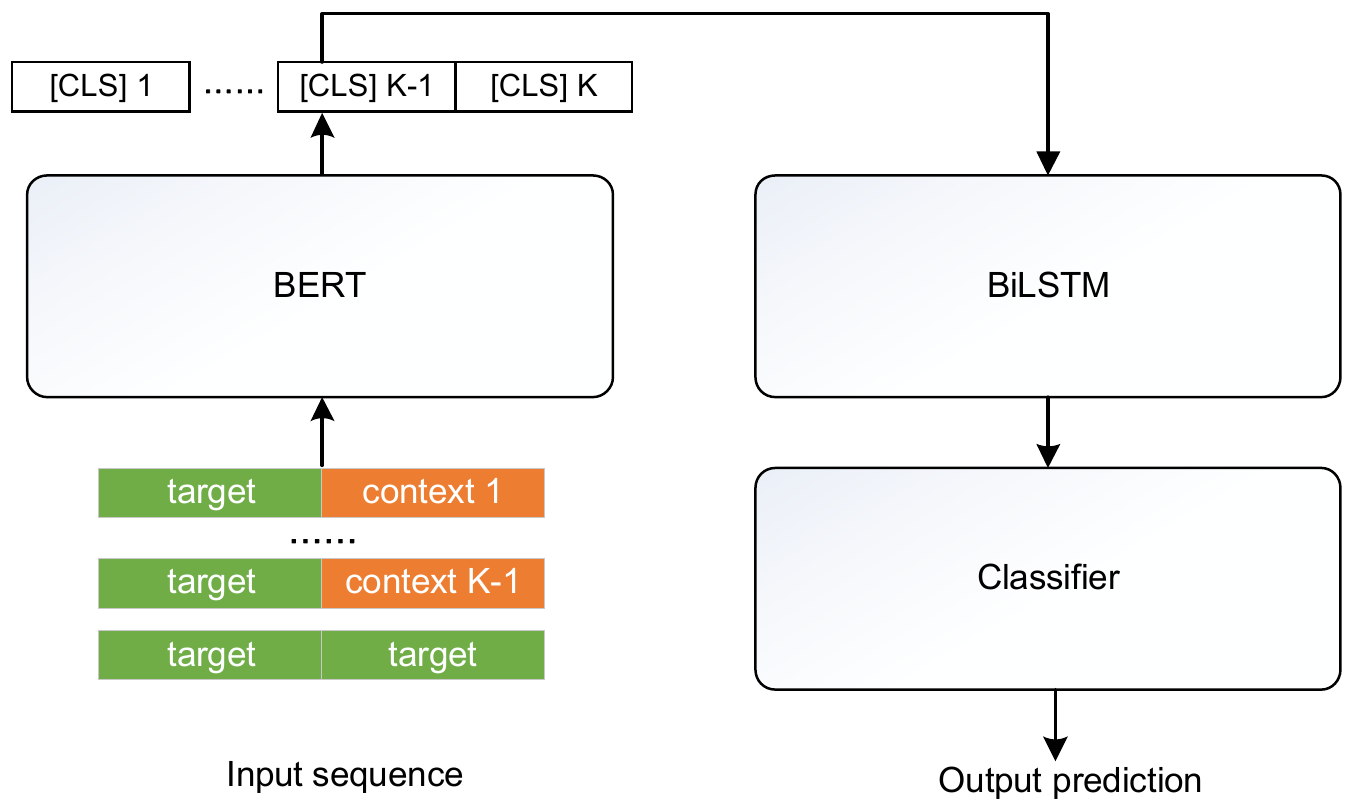}
      \caption{An overview of our BERT-based model for subtask 4.}
      \label{fig3}
    \end{figure}

    In this paper, we propose a model based on BERT whose overview is shown in Figure~\ref{fig3}.
    To detect whether every two messages belong to the same conversation, we should make each message aware of its context.
    We name a message \emph{target message}, and name its previous messages \emph{context messages}.
    Here, we heuristically consider only the nearest \emph{K} context messages of each target message, resulting in a balance between the performance and computation.

    First, each input sequence is constructed by concatenating the target message with its context messages and itself as well.
    It is noticeable that concatenating the target message with itself is designed in case that the target message is not in a conversation with any context messages\footnote{In the following part of this paper, the context messages includes the target message itself.}.
    These input sequences are first encoded by the pre-trained BERT model to obtain the sequence embeddings represented by the \texttt{[CLS]} token.
    Then, a single-layer BiLSTM is employed on top of the output of BERT in order to capture the semantics across different messages.
    We denote the outputs of the BiLSTM as $\textbf{m}^t \in \mathbb{R}^{H}$ for concatenation of the target message with itself, and $\textbf{M}^{c} \in \mathbb{R}^{K\times H}$ for concatenation of the target message with its context messages, where \emph{K} denotes the nearest \emph{K} context messages of each target message, and \emph{H} denotes the dimension of outputs of BERT.

    Furthermore, in order to model the high-order interactions between the target message and its context messages, we compute the differences and element-wise products between them.
    We duplicate the target message to obtain $\textbf{M}^{t} \in \mathbb{R}^{K\times H}$ and concatenate them as follows:
    \begin{align}
      & \textbf{M} = [\textbf{M}^{t}, \textbf{M}^{c}, \textbf{M}^{t} \odot \textbf{M}^{c}, \textbf{M}^{t} - \textbf{M}^{c}], \\
      & prediction = \textbf{Tanh}(\textbf{M} \cdot \textbf{W}^\top + \textbf{b}),
    \end{align}
    where $\textbf{W} \in \mathbb{R}^{H}$ and $\textbf{b} \in \mathbb{R}$ are parameters estimated during the training process.
    $prediction \in \mathbb{R}^{K}$ denotes the similarity scores calculated between the target message and its context messages.
    Here, we select the one obtaining the highest score with the target message, indicating which context message or none of them is in the same conversation with the target massage.

    Finally, three ensemble strategies are employed to further improve the performance as follows.
    \begin{itemize}
      \item \textbf{Model-AVG}. The final ensemble model is initialized by averaging the weights of several single models with identical architectures and different random initializations.
      \item \textbf{Probability-AVG}. Similarly, prediction probabilities for each sample are averaged across different models.
      \item \textbf{Vote-AVG}. We employ several models to make vote predictions. The context message which is voted most is considered as our final prediction in the same conversation with the target message.
    \end{itemize}


\section{Experiments}

  \subsection{Datasets}

    \begin{table}[!hbt]
      \centering
      \begin{tabular}{c|c|ccc}
      \toprule
      \multicolumn{2}{c|}{Dataset}                      & Train & Valid & Test   \\
      \hline
      \multirow{2}*{Subtask 1}    & Ubuntu              & 225,367 & 4827  & 5529 \\
                                  & Advising            & 100,000 & 500   & 500  \\
      \hline
      \multirow{1}*{Subtask 2}    & Ubuntu              & 112262 & 9565   & 9027 \\
      \hline
      \multirow{1}*{Subtask 3}    & Advising            & 500    & 500    & 500  \\
      \hline
      \multirow{1}*{Subtask 4}    & Ubuntu              & 153    & 10     & 10   \\
      \bottomrule
      \end{tabular}
      \caption{Statistics of the datasets that our models were tested on.}
      \label{tab2}
    \end{table}

    We tested our model on all subtasks of Track 2.
    Two datasets were provided under this challenge, one on the Ubuntu IRC help channel, and the other on the course recommendation between the advisor and student.
    Some statistics of these datasets were shown in Table~\ref{tab2}.

  \subsection{Evaluation Metrics}

    For subtask 1 and subtask 2, each model was tasked with selecting the $k$ best-matched responses from $n$ available candidates for the given conversation context, and we calculated the recall of the true positive replies among the $k$ selected responses, denoted as Recall@k, as the main evaluation metric.
    In addition to Recall@k, we considered the mean reciprocal rank (MRR).
    Finally, the average of Recall@10 and MRR was considered as the final metric.

    For subtask 3, the accuracy of whole dialogue prediction was considered as the main metric.
    In addition, precision, recall and F-1 value were also evaluated for reference.

    For subtask 4, five clustering metrics were adopted for evaluation: Scaled Variation of information (VI), adjusted rand index (ARI), $F_{1}$ score (F1), recall and precision.

  \subsection{Training Details}

    For subtask 1, the large version of BERT was employed.
    The Adam method \cite{DBLP:journals/corr/KingmaB14} was employed for optimization.
    The initial learning rate was set to 1e-5 and was linearly decayed by L2 weight decay.
    \texttt{gelu} activation \cite{DBLP:journals/corr/HendrycksG16} was employed.
    The maximum sequence length of the concatenation of a context-response pair was set to 320.
    The training batch size was set to 32.
    The maximum number of training epochs was set to 30.
    The dropout \cite{DBLP:journals/jmlr/SrivastavaHKSS14} probability of 0.1 is applied on all layers.
    The candidate pool may not contain the correct response, so we need to choose a threshold.
    When the probability of positive labels was smaller than the threshold, we predicted that candidate pool did not contain the correct response.
    The threshold was selected from the range [0.6, 0.65, .., 0.95] based on the validation set and was set to 0.95 finally.
    We used the validation set to set the stop condition to select the best model for testing.

    For subtask 2, the base version of BERT was employed because the large version could not provide further improvement.
    The initial learning rate was set to 2e-5.
    The maximum sequence length of the concatenation of a context-response pair was set to 512.
    The training batch size was set to 25.
    The maximum number of training epochs was set to 8.
    The threshold to decide whether the candidate pool contains the correct response was set to 0.95.

    For subtask 3, the word representations were 300-dimensional GloVe embeddings \cite{DBLP:conf/emnlp/PenningtonSM14}, the 100-dimensional embeddings estimated on the training set using the Word2Vec algorithm \cite{DBLP:conf/nips/MikolovSCCD13} and the 150-dimensional character-level embeddings with window sizes of \{3, 4,  5\}, each consisting of 50 filters.
    The word embeddings were not updated during training.
    All hidden states of the LSTM had 200 dimensions.
    The MLP at the prediction had 256 hidden units with ReLU \cite{DBLP:conf/icml/NairH10} activation.
    Dropout with a rate of 0.2 was applied to the word embeddings and all hidden layers.
    The maximum utterance length, maximum number of utterances in a context were set to 30 and 26 respectively.
    Zeros were padded if the number of utterances in a context was less than 26.
    Otherwise, we kept the last 26 utterances in the context.
    The Adam method was employed for optimization with a batch size of 200.
    The learning rate was initialized as 0.001 and was exponentially decayed by 0.96 every 5000 steps.
    The weight of loss for \emph{Accept} or \emph{Reject} is enlarged to twice the original, and that for \emph{No Decision Yet} keeps the original.

    For subtask 4, we used the base version of BERT, because no further improvement could be achieved by its large version.
    The initial learning rate was set to 2e-5.
    The maximum sequence length were set to 100.
    The batch size was set to 4 .
    The max number of messages which were considered as the context of the target message was set to 50.
    Messages after the target message were also taken into consideration, but no further improvement was achieved.
    The hidden size of the BiLSTM module were set to 384 to make the concatenated output equal to 768 which was the same size as the output of BERT.
    The heuristic classifier had 3072 hidden units.
    For ensemble strategies, different strategies require different numbers of models to achieve the best result.
    For Model-AVG, the number was set to 2.
    For both Probability-AVG and Vote-AVG, the number was set to 8.


  \subsection{Experimental Results}

    \begin{table}
      \centering
      \begin{tabular}{llcc}
      \toprule
       Subtask                   & Measure   &  Ubuntu             &  Advising  \\
      \midrule
       \multirow{4}*{Subtask 1}  & Recall@1  &  0.649              &  0.224     \\
                                 & Recall@5  &  0.904              &  0.526     \\
                                 & Recall@10 &  0.949              &  0.676     \\
                                 & MRR       &  0.760              &  0.374     \\
      \midrule
       \multirow{4}*{Subtask 2}  & Recall@1  &  0.506              &  \multirow{4}*{NA}     \\
                                 & Recall@5  &  0.755              &       \\
                                 & Recall@10 &  0.834              &       \\
                                 & MRR       &  0.621              &       \\
      \midrule
       \multirow{4}*{Subtask 3}  & Accuracy  &  \multirow{4}*{NA}  &  0.802     \\
                                 & Precision &                     &  0.832     \\
                                 & Recall    &                     &  0.802     \\
                                 & F-1       &                     &  0.817     \\
      \midrule
       \multirow{6}*{Subtask 4}  & Precision &  0.443              &  \multirow{6}*{NA}     \\
                                 & Recall    &  0.496              &       \\
                                 & F-score   &  0.468              &       \\
                                 & VI        &  0.933              &       \\
                                 & Rand      &  0.752              &       \\
                                 & AMI       &  0.865              &       \\

      \bottomrule
      \end{tabular}
      \caption{The submission results on the hidden test sets for the Track 2 of DSTC 8 challenge. NA - not applicable.}
      \label{tab3}
    \end{table}

    Table~\ref{tab3} presents the evaluation results of our methods on the four subtasks.
    We tuned our single models on the validation set and submitted the final results using ensemble models.
    The ensemble models were built by averaging the outputs of five single models with identical architectures and different random initializations.
    Finally, our results ranked fourth in subtask 1, third in subtask 2, and first in subtask 3 and subtask 4 respectively.

\section{Analysis}

    \begin{table}
      \centering
      \setlength{\tabcolsep}{1.5mm}{
      \begin{tabular}{lcccc}
      \toprule
                   &  Recall@1  &  Recall@5  &  Recall@10  &  MRR   \\
      \midrule
       Ours        &  0.638     &  0.895     &  0.938      &  0.749 \\
       ~-Pre-train &  0.622     &  0.881     &  0.909      &  0.733 \\
       ~~-Switch   &  0.616     &  0.877     &  0.902      &  0.728 \\
      \bottomrule
      \end{tabular}}
      \caption{Ablation results for a single model on the validation set of Ubuntu dataset in subtask 1.}
      \label{tab4}
    \end{table}

    \begin{table}
      \centering
      \setlength{\tabcolsep}{1.2mm}{
      \begin{tabular}{lcccc}
      \toprule
                       &  Recall@1  &  Recall@5  &  Recall@10  &  MRR   \\
      \midrule
       Ours            &  0.477     &  0.728     &  0.810      &  0.594 \\
       ~-Switch        &  0.452     &  0.713     &  0.799      &  0.573 \\
       ~~-Pre-train    &  0.436     &  0.701     &  0.790      &  0.559 \\
       ~~~-Disentangle &  0.258     &  0.393     &  0.458      &  0.335 \\
      \bottomrule
      \end{tabular}}
      \caption{Ablation results for a single model on the validation set of Ubuntu dataset in subtask 2.}
      \label{tab5}
    \end{table}

    \begin{table}
      \centering
      \begin{tabular}{lc}
      \toprule
                       &  Accuracy   \\
      \midrule
       Ours            &  0.868      \\
       ~-Paraphrase    &  0.852      \\
       ~~-Weight loss  &  0.846      \\
      \bottomrule
      \end{tabular}
      \caption{Ablation results for a single model on the validation set of Advising dataset in subtask 3.}
      \label{tab6}
    \end{table}

    \begin{table}
      \centering
      \setlength{\tabcolsep}{1.2mm}{
      \begin{tabular}{lccccc}
      \toprule
                       &  1-Scaled VI  &  ARI  & F1    & recall & precision  \\
      \midrule
       Ours            &  0.947        & 0.841 & 0.463 & 0.502  & 0.482     \\
       ~-features      &  0.937        & 0.813 & 0.482 & 0.497  & 0.490      \\
       baseline        &  0.921        & 0.742 & 0.405 & 0.412  & 0.409      \\
      \bottomrule
      \end{tabular}}
      \caption{Ablation results for a single model on the validation set of subtask 4.}
      \label{tab7}
    \end{table}

    \begin{table}
      \centering
      \setlength{\tabcolsep}{0.7mm}{
      \begin{tabular}{lccccc}
      \toprule
                          &  1-Scaled VI    &  ARI           & F1             & recall         & precision  \\
      \midrule
       Model-AVG          &  0.939          & 0.811          & 0.472          & 0.520          & 0.495     \\
       Probability-AVG    &  \textbf{0.947} & \textbf{0.831} & \textbf{0.521} & 0.547          & 0.534      \\
       Vote-AVG           &  0.941          & 0.783          & 0.519          & \textbf{0.552} & \textbf{0.535}      \\
      \bottomrule
      \end{tabular}}
      \caption{Results for a ensemble model on the validation set of subtask 4.}
      \label{tab8}
    \end{table}

    To demonstrate the importance of each component in our proposed model, various parts of the architecture were ablated, and the results were reported on the validation set on each subtask, as shown in Table~\ref{tab4}, Table~\ref{tab5}, Table~\ref{tab6}, Table~\ref{tab7} and Table~\ref{tab8}.

    From Table~\ref{tab4}, we can see that both the pre-training process and switch embeddings contribute to our final model.
    Without the pre-training process, the metric of Recall@1 drops a large margin 1.6\%, which shows that the external data given by the DSTC 8 package does improve our model further when we use pre-training.
    Furthermore, the performance continues to drop 0.6\% in terms of Recall@1 by ablating the switch embeddings, which shows the effectiveness of utilizing the information of speaker turns.

    Similarly, the pre-training process and switch embeddings also benefit subtask 2 as shown in Table~\ref{tab5}.
    In addition, we ablated the disentanglement strategy and truncated the sequence to the max sequence length of BERT by selecting the head or tail part.
    The performance drops sharply which shows the effectiveness of the disentanglement strategy.

    Table~\ref{tab6} shows that enriching the training data with the help of the paraphrase and the weighted loss function are both effective.

    From the result shown in Table~\ref{tab7}, we can see that our model has outperform the baseline given by DSTC 8 contest in all the five evaluation metrics.
    After we combine feature such as user and time etc., our model could achieve the further improvement, which indicates that manual feature still could capture some information that BERT could not capture.
    Even though we use the state-of-art model BERT with pre-training method, our model could only reach 46.3\% F1 score, which indicates that the disentangling problem is still a hard problem to solve.

    Table~\ref{tab8} shows that, the Probability-AVG and Vote-AVG strategies could reach better performance compared with Model-AVG.
    Probability-AVG performs better on 1-Scaled VI, ARI and F1 metrics, and Votes performs better on recall and precision metrics.

\section{Conclusion}

  This paper describes our systems that are evaluated on all subtasks of Track 2 of DSTC 8.
  Pre-trained attention-based network for multi-turn dialogue systems are designed for each subtask according to different evaluation dimensions.
  In the released evaluation results of Track 2 of DSTC 8, our proposed models ranked fourth in subtask 1, third in subtask 2, and first in subtask 3 and subtask 4 respectively.
  Investigating other strategies for better employing pre-trained language models for multi-turn dialogue will be a part of our future work.

\bibliography{mybib}

\begin{thebibliography}{}

\bibitem[\protect\citeauthoryear{Aoki \bgroup et al\mbox.\egroup
  }{2006}]{DBLP:journals/corr/abs-cs-0608083}
Aoki, P.~M.; Szymanski, M.~H.; Plurkowski, L.~D.; Thornton, J.~D.; Woodruff,
  A.; and Yi, W.
\newblock 2006.
\newblock Where's the "party" in "multi-party"? analyzing the structure of
  small-group sociable talk.
\newblock {\em CoRR} abs/cs/0608083.

\bibitem[\protect\citeauthoryear{Bowman \bgroup et al\mbox.\egroup
  }{2015}]{DBLP:conf/emnlp/BowmanAPM15}
Bowman, S.~R.; Angeli, G.; Potts, C.; and Manning, C.~D.
\newblock 2015.
\newblock A large annotated corpus for learning natural language inference.
\newblock In {\em Proceedings of the 2015 Conference on Empirical Methods in
  Natural Language Processing, {EMNLP} 2015, Lisbon, Portugal, September 17-21,
  2015},  632--642.

\bibitem[\protect\citeauthoryear{Devlin \bgroup et al\mbox.\egroup
  }{2019}]{DBLP:conf/naacl/DevlinCLT19}
Devlin, J.; Chang, M.; Lee, K.; and Toutanova, K.
\newblock 2019.
\newblock {BERT:} pre-training of deep bidirectional transformers for language
  understanding.
\newblock In {\em Proceedings of the 2019 Conference of the NAACL-HLT 2019,
  Minneapolis, MN, USA, June 2-7, 2019, Volume 1 (Long and Short Papers)},
  4171--4186.

\bibitem[\protect\citeauthoryear{Dong and
  Huang}{2018}]{DBLP:journals/corr/abs-1802-02614}
Dong, J., and Huang, J.
\newblock 2018.
\newblock Enhance word representation for out-of-vocabulary on ubuntu dialogue
  corpus.
\newblock {\em CoRR} abs/1802.02614.

\bibitem[\protect\citeauthoryear{Gu \bgroup et al\mbox.\egroup
  }{2019}]{gu-etal-2019-dually}
Gu, J.-C.; Ling, Z.-H.; Zhu, X.; and Liu, Q.
\newblock 2019.
\newblock Dually interactive matching network for personalized response
  selection in retrieval-based chatbots.
\newblock In {\em Proceedings of the 2019 Conference on Empirical Methods in
  Natural Language Processing and the 9th International Joint Conference on
  Natural Language Processing (EMNLP-IJCNLP)},  1845--1854.
\newblock Hong Kong, China: Association for Computational Linguistics.

\bibitem[\protect\citeauthoryear{Gu, Ling, and
  Liu}{2019a}]{Gu:2019:IMN:3357384.3358140}
Gu, J.-C.; Ling, Z.-H.; and Liu, Q.
\newblock 2019a.
\newblock Interactive matching network for multi-turn response selection in
  retrieval-based chatbots.
\newblock In {\em Proceedings of the 28th ACM International Conference on
  Information and Knowledge Management}, CIKM '19,  2321--2324.
\newblock ACM.

\bibitem[\protect\citeauthoryear{Gu, Ling, and Liu}{2019b}]{gu2019utterance}
Gu, J.-C.; Ling, Z.-H.; and Liu, Q.
\newblock 2019b.
\newblock Utterance-to-utterance interactive matching network for multi-turn
  response selection in retrieval-based chatbots.
\newblock {\em IEEE/ACM Transactions on Audio, Speech, and Language Processing}
  28:369--379.

\bibitem[\protect\citeauthoryear{Gunasekara \bgroup et al\mbox.\egroup
  }{2019}]{gunasekara2019dstc7}
Gunasekara, C.; Kummerfeld, J.~K.; Polymenakos, L.; and Lasecki, W.
\newblock 2019.
\newblock Dstc7 task 1: Noetic end-to-end response selection.
\newblock In {\em Proceedings of the First Workshop on NLP for Conversational
  AI},  60--67.

\bibitem[\protect\citeauthoryear{Hendrycks and
  Gimpel}{2016}]{DBLP:journals/corr/HendrycksG16}
Hendrycks, D., and Gimpel, K.
\newblock 2016.
\newblock Bridging nonlinearities and stochastic regularizers with gaussian
  error linear units.
\newblock {\em CoRR} abs/1606.08415.

\bibitem[\protect\citeauthoryear{Hochreiter and
  Schmidhuber}{1997}]{DBLP:journals/neco/HochreiterS97}
Hochreiter, S., and Schmidhuber, J.
\newblock 1997.
\newblock Long short-term memory.
\newblock {\em Neural Computation} 9(8):1735--1780.

\bibitem[\protect\citeauthoryear{Ji, Lu, and
  Li}{2014}]{DBLP:journals/corr/JiLL14}
Ji, Z.; Lu, Z.; and Li, H.
\newblock 2014.
\newblock An information retrieval approach to short text conversation.
\newblock {\em CoRR} abs/1408.6988.

\bibitem[\protect\citeauthoryear{Jiang \bgroup et al\mbox.\egroup
  }{2018}]{jiang2018learning}
Jiang, J.-Y.; Chen, F.; Chen, Y.-Y.; and Wang, W.
\newblock 2018.
\newblock Learning to disentangle interleaved conversational threads with a
  siamese hierarchical network and similarity ranking.
\newblock In {\em Proceedings of the 2018 Conference of the NAACL-HLT, Volume 1
  (Long Papers)},  1812--1822.

\bibitem[\protect\citeauthoryear{Kim \bgroup et al\mbox.\egroup }{2019}]{DSTC8}
Kim, S.; Galley, M.; Gunasekara, C.; Lee, S.; Atkinson, A.; Peng, B.; Schulz,
  H.; Gao, J.; Li, J.; Adada, M.; et~al.
\newblock 2019.
\newblock The eighth dialog system technology challenge.
\newblock {\em arXiv preprint arXiv:1911.06394}.

\bibitem[\protect\citeauthoryear{Kingma and
  Ba}{2014}]{DBLP:journals/corr/KingmaB14}
Kingma, D.~P., and Ba, J.
\newblock 2014.
\newblock Adam: {A} method for stochastic optimization.
\newblock {\em CoRR} abs/1412.6980.

\bibitem[\protect\citeauthoryear{Kummerfeld \bgroup et al\mbox.\egroup
  }{2019}]{kummerfeld2019large}
Kummerfeld, J.~K.; Gouravajhala, S.~R.; Peper, J.~J.; Athreya, V.; Gunasekara,
  C.; Ganhotra, J.; Patel, S.~S.; Polymenakos, L.~C.; and Lasecki, W.
\newblock 2019.
\newblock A large-scale corpus for conversation disentanglement.
\newblock In {\em Proceedings of the 57th Annual Meeting of the Association for
  Computational Linguistics},  3846--3856.

\bibitem[\protect\citeauthoryear{Lowe \bgroup et al\mbox.\egroup
  }{2015}]{DBLP:conf/sigdial/LowePSP15}
Lowe, R.; Pow, N.; Serban, I.; and Pineau, J.
\newblock 2015.
\newblock The ubuntu dialogue corpus: {A} large dataset for research in
  unstructured multi-turn dialogue systems.
\newblock In {\em Proceedings of the {SIGDIAL} 2015 Conference, The 16th Annual
  Meeting of the Special Interest Group on Discourse and Dialogue, 2-4
  September 2015, Prague, Czech Republic},  285--294.

\bibitem[\protect\citeauthoryear{Lowe \bgroup et al\mbox.\egroup
  }{2017}]{DBLP:journals/dad/LowePSCLP17}
Lowe, R.~T.; Pow, N.; Serban, I.~V.; Charlin, L.; Liu, C.; and Pineau, J.
\newblock 2017.
\newblock Training end-to-end dialogue systems with the ubuntu dialogue corpus.
\newblock {\em D{\&}D} 8(1):31--65.

\bibitem[\protect\citeauthoryear{Mehri and Carenini}{2017}]{mehri2017chat}
Mehri, S., and Carenini, G.
\newblock 2017.
\newblock Chat disentanglement: Identifying semantic reply relationships with
  random forests and recurrent neural networks.
\newblock In {\em Proceedings of the Eighth International Joint Conference on
  Natural Language Processing (Volume 1: Long Papers)},  615--623.

\bibitem[\protect\citeauthoryear{Mikolov \bgroup et al\mbox.\egroup
  }{2013}]{DBLP:conf/nips/MikolovSCCD13}
Mikolov, T.; Sutskever, I.; Chen, K.; Corrado, G.~S.; and Dean, J.
\newblock 2013.
\newblock Distributed representations of words and phrases and their
  compositionality.
\newblock In {\em Advances in neural information processing systems},
  3111--3119.

\bibitem[\protect\citeauthoryear{Nair and
  Hinton}{2010}]{DBLP:conf/icml/NairH10}
Nair, V., and Hinton, G.~E.
\newblock 2010.
\newblock Rectified linear units improve restricted boltzmann machines.
\newblock In {\em Proceedings of the 27th International Conference on Machine
  Learning (ICML-10), June 21-24, 2010, Haifa, Israel},  807--814.

\bibitem[\protect\citeauthoryear{Pennington, Socher, and
  Manning}{2014}]{DBLP:conf/emnlp/PenningtonSM14}
Pennington, J.; Socher, R.; and Manning, C.~D.
\newblock 2014.
\newblock Glove: Global vectors for word representation.
\newblock In {\em Proceedings of the 2014 Conference on Empirical Methods in
  Natural Language Processing, {EMNLP} 2014, October 25-29, 2014, Doha, Qatar,
  {A} meeting of SIGDAT, a Special Interest Group of the {ACL}},  1532--1543.

\bibitem[\protect\citeauthoryear{Peters \bgroup et al\mbox.\egroup
  }{2018}]{DBLP:conf/naacl/PetersNIGCLZ18}
Peters, M.~E.; Neumann, M.; Iyyer, M.; Gardner, M.; Clark, C.; Lee, K.; and
  Zettlemoyer, L.
\newblock 2018.
\newblock Deep contextualized word representations.
\newblock In {\em Proceedings of the 2018 Conference of the NAACL-HLT 2018, New
  Orleans, Louisiana, USA, June 1-6, 2018, Volume 1 (Long Papers)},
  2227--2237.

\bibitem[\protect\citeauthoryear{Rajpurkar \bgroup et al\mbox.\egroup
  }{2016}]{DBLP:conf/emnlp/RajpurkarZLL16}
Rajpurkar, P.; Zhang, J.; Lopyrev, K.; and Liang, P.
\newblock 2016.
\newblock Squad: 100, 000+ questions for machine comprehension of text.
\newblock In {\em Proceedings of the 2016 Conference on Empirical Methods in
  Natural Language Processing, {EMNLP} 2016, Austin, Texas, USA, November 1-4,
  2016},  2383--2392.

\bibitem[\protect\citeauthoryear{Riou, Salim, and
  Hernandez}{2015}]{riou:hal-01698147}
Riou, M.; Salim, S.; and Hernandez, N.
\newblock 2015.
\newblock {Using discursive information to disentangle French language chat}.
\newblock In {\em {2nd Workshop on Natural Language Processing for
  Computer-Mediated Communication (NLP4CMC 2015) / Social Media at GSCL
  Conference 2015}},  23--27.

\bibitem[\protect\citeauthoryear{Serban \bgroup et al\mbox.\egroup
  }{2016}]{DBLP:conf/aaai/SerbanSBCP16}
Serban, I.~V.; Sordoni, A.; Bengio, Y.; Courville, A.~C.; and Pineau, J.
\newblock 2016.
\newblock Building end-to-end dialogue systems using generative hierarchical
  neural network models.
\newblock In {\em Proceedings of the Thirtieth {AAAI} Conference on Artificial
  Intelligence, February 12-17, 2016, Phoenix, Arizona, {USA.}},  3776--3784.

\bibitem[\protect\citeauthoryear{Shang, Lu, and
  Li}{2015}]{DBLP:conf/acl/ShangLL15}
Shang, L.; Lu, Z.; and Li, H.
\newblock 2015.
\newblock Neural responding machine for short-text conversation.
\newblock In {\em Proceedings of the 53rd Annual Meeting of the Association for
  Computational Linguistics and the 7th International Joint Conference on
  Natural Language Processing of the Asian Federation of Natural Language
  Processing, {ACL} 2015, July 26-31, 2015, Beijing, China, Volume 1: Long
  Papers},  1577--1586.

\bibitem[\protect\citeauthoryear{Srivastava \bgroup et al\mbox.\egroup
  }{2014}]{DBLP:journals/jmlr/SrivastavaHKSS14}
Srivastava, N.; Hinton, G.~E.; Krizhevsky, A.; Sutskever, I.; and
  Salakhutdinov, R.
\newblock 2014.
\newblock Dropout: a simple way to prevent neural networks from overfitting.
\newblock {\em Journal of Machine Learning Research} 15(1):1929--1958.

\bibitem[\protect\citeauthoryear{Tao \bgroup et al\mbox.\egroup
  }{2019a}]{DBLP:conf/wsdm/TaoWXHZY19}
Tao, C.; Wu, W.; Xu, C.; Hu, W.; Zhao, D.; and Yan, R.
\newblock 2019a.
\newblock Multi-representation fusion network for multi-turn response selection
  in retrieval-based chatbots.
\newblock In {\em Proceedings of the Twelfth {ACM} International Conference on
  Web Search and Data Mining, {WSDM} 2019, Melbourne, VIC, Australia, February
  11-15, 2019},  267--275.

\bibitem[\protect\citeauthoryear{Tao \bgroup et al\mbox.\egroup
  }{2019b}]{DBLP:conf/acl/TaoWXHZY19}
Tao, C.; Wu, W.; Xu, C.; Hu, W.; Zhao, D.; and Yan, R.
\newblock 2019b.
\newblock One time of interaction may not be enough: Go deep with an
  interaction-over-interaction network for response selection in dialogues.
\newblock In {\em Proceedings of the 57th Conference of the Association for
  Computational Linguistics, {ACL} 2019, Florence, Italy, July 28- August 2,
  2019, Volume 1: Long Papers},  1--11.

\bibitem[\protect\citeauthoryear{Vaswani \bgroup et al\mbox.\egroup
  }{2017}]{DBLP:conf/nips/VaswaniSPUJGKP17}
Vaswani, A.; Shazeer, N.; Parmar, N.; Uszkoreit, J.; Jones, L.; Gomez, A.~N.;
  Kaiser, L.; and Polosukhin, I.
\newblock 2017.
\newblock Attention is all you need.
\newblock In {\em Advances in Neural Information Processing Systems 30: Annual
  Conference on Neural Information Processing Systems 2017, 4-9 December 2017,
  Long Beach, CA, {USA}},  5998--6008.

\bibitem[\protect\citeauthoryear{Wang \bgroup et al\mbox.\egroup
  }{2013}]{DBLP:conf/emnlp/WangLLC13}
Wang, H.; Lu, Z.; Li, H.; and Chen, E.
\newblock 2013.
\newblock A dataset for research on short-text conversations.
\newblock In {\em Proceedings of the 2013 Conference on Empirical Methods in
  Natural Language Processing, {EMNLP} 2013},  935--945.

\bibitem[\protect\citeauthoryear{Wu \bgroup et al\mbox.\egroup
  }{2017}]{DBLP:conf/acl/WuWXZL17}
Wu, Y.; Wu, W.; Xing, C.; Zhou, M.; and Li, Z.
\newblock 2017.
\newblock Sequential matching network: {A} new architecture for multi-turn
  response selection in retrieval-based chatbots.
\newblock In {\em Proceedings of the 55th Annual Meeting of the Association for
  Computational Linguistics, {ACL} 2017},  496--505.

\bibitem[\protect\citeauthoryear{Yang \bgroup et al\mbox.\egroup
  }{2019}]{DBLP:journals/corr/abs-1906-08237}
Yang, Z.; Dai, Z.; Yang, Y.; Carbonell, J.~G.; Salakhutdinov, R.; and Le, Q.~V.
\newblock 2019.
\newblock Xlnet: Generalized autoregressive pretraining for language
  understanding.
\newblock {\em CoRR} abs/1906.08237.

\bibitem[\protect\citeauthoryear{Zhang \bgroup et al\mbox.\egroup
  }{2018}]{DBLP:conf/coling/ZhangLZZL18}
Zhang, Z.; Li, J.; Zhu, P.; Zhao, H.; and Liu, G.
\newblock 2018.
\newblock Modeling multi-turn conversation with deep utterance aggregation.
\newblock In {\em Proceedings of the 27th International Conference on
  Computational Linguistics, {COLING} 2018},  3740--3752.

\bibitem[\protect\citeauthoryear{Zhou \bgroup et al\mbox.\egroup
  }{2018}]{DBLP:conf/acl/WuLCZDYZL18}
Zhou, X.; Li, L.; Dong, D.; Liu, Y.; Chen, Y.; Zhao, W.~X.; Yu, D.; and Wu, H.
\newblock 2018.
\newblock Multi-turn response selection for chatbots with deep attention
  matching network.
\newblock In {\em Proceedings of the 56th Annual Meeting of the Association for
  Computational Linguistics, {ACL} 2018},  1118--1127.

\end{thebibliography}
\bibliographystyle{aaai}

\end{document}